\documentclass[10pt,twocolumn,letterpaper]{article}

\usepackage{cvpr}      % To produce the REVIEW version
\usepackage{graphicx}
\usepackage{amsmath}
\usepackage{booktabs}
\usepackage{amsfonts}
\usepackage{multirow}
\usepackage{utfsym}
\usepackage{ulem}
\usepackage{float}

\newcommand{\w}{{\bf t}}

\newcommand{\iii}{{\bf i}}

\newcommand{\y}{{\bf y}}
\newcommand{\p}{{\bf p}}

\newcommand{\EE}{\mathbb{E}}

\usepackage[pagebackref,breaklinks,colorlinks]{hyperref}

\usepackage[capitalize]{cleveref}
\crefname{section}{Sec.}{Secs.}
\Crefname{section}{Section}{Sections}
\Crefname{table}{Table}{Tables}
\crefname{table}{Tab.}{Tabs.}

\def\cvprPaperID{*****} % *** Enter the CVPR Paper ID here
\def\confName{CVPR}
\def\confYear{2022}

\begin{document}

\title{The Solution for the CVPR2023 NICE Image Captioning Challenge}

\author{
Xiangyu Wu,  
YiGao,  
Hailiang Zhang,  
YangYang,  
Weili Guo,  
Jianfeng Lu  
\\Nanjing University of Science and Technology
}

\maketitle

%%%%%%%%% ABSTRACT
\begin{abstract}
In this paper, we present our solution to the New Frontiers for Zero-shot Image Captioning Challenge. Different from the traditional image captioning datasets, this challenge includes a larger new variety of visual concepts from many domains (such as COVID-19) as well as various image types (photographs, illustrations, graphics). For the data level, we collect external training data from Laion5B, a large-scale CLIP-filtered image-text dataset. For the model level, we use OFA, a large-scale visual-language pre-training model based on handcrafted templates, to perform the image captioning task. In addition, we introduce contrastive learning to align image-text pairs to learn new visual concepts in the pre-training stage. Then, we propose a similarity-bucket strategy and incorporate this strategy into the template to force the model to generate higher quality and more matching captions. Finally, by retrieval-augmented strategy, we construct a contentrich template, containing the most relevant top-k captions from other image-text pairs, to guide the model in generating semantic-rich captions. Our method ranks first on the leaderboard, achieving 105.17 and 325.72 Cider-Score in the validation and test phase, respectively.
\end{abstract}

% \begin{keywords}
% multi-head attention, multi-modal, visual question answering, channel attention
% \end{keywords}

%%%%%%%%% BODY TEXT
\section{Introduction}
\label{sec:intro}

Zero-shot image captioning~\cite{Fei_Yan_Wang_Tian, Chen_Guo_Yi_Li_Elhoseiny_2021, NOA,Paper-1} requires joint modeling for vision and language, which aims to generate a concise textual summary for a given image. However, in real-world scenarios, high-quality human-annotated data are always difficult to obtain. Therefore, it is crucial and feasible that learn aligned vision and language representations from large-scale web-crewed data and transfer knowledge from pre-training models to downstream tasks.

\begin{figure}[t]
	\centering
	\includegraphics[width=\linewidth]{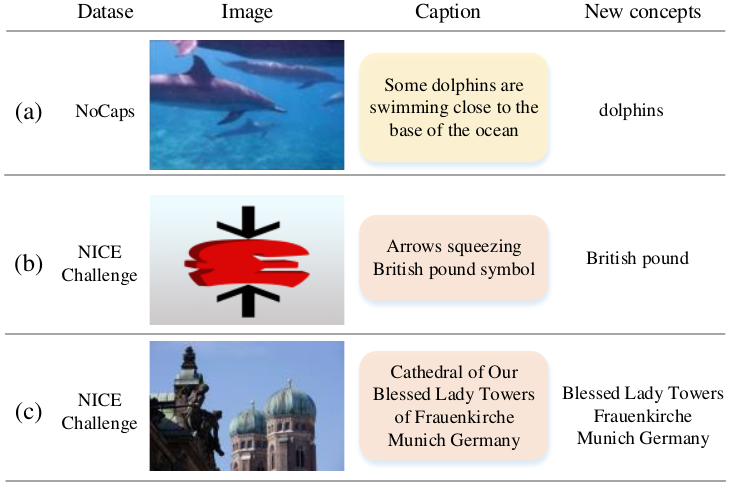}
	\caption{(a): NoCaps dataset, which always includes common objects such as animals, plants and furniture, etc. (b)(c): NICE Challenge dataset, which includes many novel visual concepts and various image types, such as famous historic, cultural and graphics,etc.}\label{fig: dataset}
\end{figure}

NoCaps~\cite{nocaps} is an extensive attended dataset for zero-shot image captioning, which contains nearly 400 object classes seen in test images. However, different from NoCaps, the competition dataset includes a larger
variety of novel visual concepts as well as various image types ( an example is shown in Figure \ref{fig: dataset}). 

To accomplish this task, the models need to broadly understand vision-language relations and simultaneously learn how to combine language components for a new concept of image. Therefore, we use OFA~\cite{ofa}, a large-scale Task-Agnostic and Modality-Agnostic pre-training framework, as our base model, and Laion-5B~\cite{laion}, a dataset of 585 billion CLIP-filtered image-text pairs, as the main source of our base dataset. 

\begin{figure*}[!h]
	\centering
	\includegraphics[width=\linewidth]{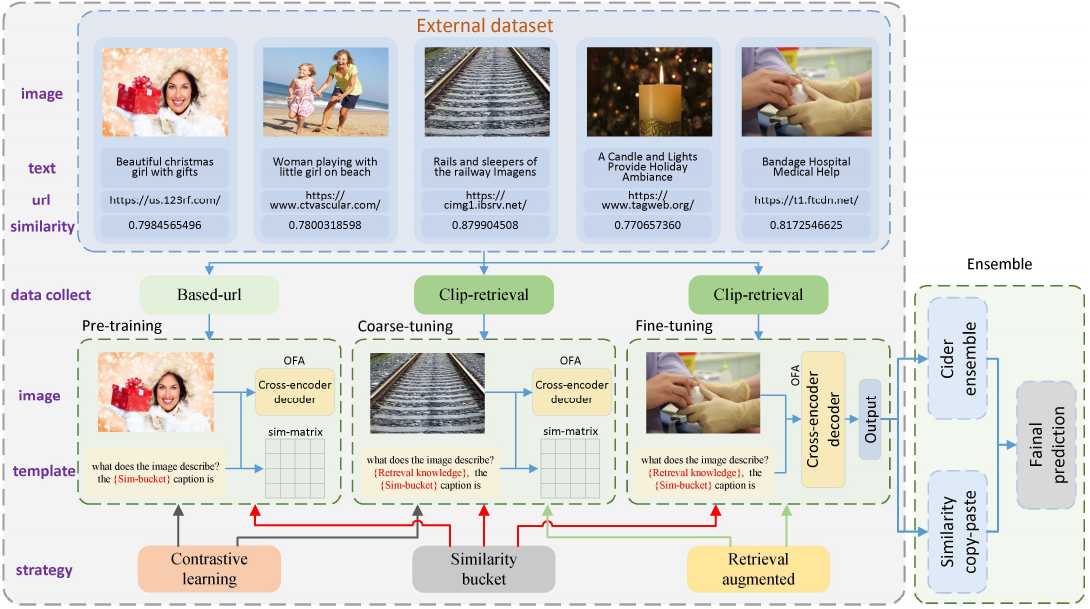}
	\caption{Overall Architecture. Our solution consists of four main stages, which include Pre-training, Coarse-tuning, Fine-tuning, and Model-ensemble. The training data for the first three stages are all collected from the large-scale Laion-5B dataset.}\label{fig:overview}
\end{figure*}

In addition, we explore some strategies to enhance zero-shot capacity: (1) we introduce contrastive learning on the OFA model to align a wide range of visual conceptual representations. (2) we propose a similarity-bucket strategy to incorporate image-text similarity into the template and guide model to generate high-quality and more matching captions. (3) we employ a retrieval-augmented~\cite{retrieval-augmented,Paper-3,Paper-4,Paper-5} strategy to enrich the textual information of the template. As a result, our method ranks first on the leaderboard, producing 105.17 and 325.72 Cider-Score in the validation and test phase, respectively. In the remainder of this technical report, we will introduce the detailed architecture of our solution for this challenge.

%------------------------------------------------------------------------

\section{Related Work}
\label{sec:Related}
\subsection{Vision-language Pre-training}

Vision-language pre-training (VLP)~\cite{blip2, mplug,Paper-6,Paper-7} aims to improve the performance of downstream tasks by pre-training the model on large-scale image-text pairs. The single-stream models refer to models where the text and visual features are concatenated together and then fed into a single transformer block. The dual-stream models refer to models where the text and visual features are not concatenated together but sent to two different transformer blocks independently. Different from the above two architectures, the OFA~\cite{ofa} model formulates both pretraining and finetuning tasks in a unified sequence-to-sequence abstraction via handcrafted instructions to achieve Task-Agnostic. 

%-------------------------------------------------------------------------
\subsection{Image Captioning}

Zero-shot image captioning~\cite{Paper-9,Paper-10,Paper-11} aims to generate textual descriptions without human-annotated data. This involves developing algorithms and models that can analyze the visual content and generate a corresponding textual description. Image captioning has a wide range of applications, including image and video search, assistive technologies for the visually impaired, and automated content generation for social media and marketing. Feng et al.~\cite{caption-1} propose unsupervised captioning without using paired image-caption supervision. Kim et al ~\cite{caption-2} focus on learning efficiency and improving data efficiency by learning from auxiliary unpaired image-caption data.

%-------------------------------------------------------------------------
\subsection{Vision-Language Retrieval}

Vision-Language Retrieval~\cite{retrieval-1, retrieval-2, DOMFN,Paper-12} aims to learn consistent representations of different modalities, to retrieve instances of one modality based on queries from another modality. The goal of vision-language retrieval is to create more intuitive and effective ways for humans to interact with machines, such as through image and video search, automated image captioning, and visual question answering. For example, ~\cite{Fei_Yan_Wang_Tian} proposed a similarity graph reasoning module relying on a graph convolution neural
network. With the developments in Transformer-based language understanding, large-scale vision-language transformers have inspired deeper modal interaction in retrieval models.

%------------------------------------------------------------------------
\section{Methodology}
\subsection{Overall Architecture}

Figure 2 illustrates the overall architecture of our solution, which contains four components: Pre-training, Coarse-tuning, Fine-tuning, and Model-ensemble. 

Pre-training stage collects specific \textit{url} data from Laion-5B, which can align a wide range of image-text concepts and store sufficient vision-language knowledge through contrastive learning, image captioning pre-train objectives, and handcrafted templates with similarity-bucket strategy. 

Coarse-tuning stage utilizes the \textit{Clip-retrieval}~\cite{clip-retrieval} library to retrieve small-scale datasets with simple data cleaning, which can learn a large variety of vision-language novel concepts similar to the competition domain. In this stage, we introduce a retrieval-augmented strategy to retrieve abundant textual knowledge and integrate it into the template. As same as the pre-training stage, we utilize contrastive learning, image captioning, and similarity-bucket to coarse tuning.

Fine-tuning stage further compresses the retrieved dataset in the coarse-tuning stage and adds it to the competition validation dataset. By applying similarity-bucket and retrieval-augmented strategies, the zero-shot performance of the model can be effectively improved. 

Model-ensemble is the last stage, which uses similarity-copy-paste and cider-ensemble tricks to improve the generalization ability of the model on the zero-shot image captioning task.

Next, we will specifically introduce contrastive-learning, similarity-bucket, retrieval-augmented strategies, and model-ensemble tricks, as well as how to integrate these strategies into the vision-language pre-training model OFA.

\subsection{Contrastive-learning}

\textbf{Contrastive-learning} aims to learn better uni-modal representations before fusion. A similarity function is learned in which the parallel image-text pairs are assigned higher similarity scores. We take the [CLS] embeddings output by image and text encoders as joint representations for prediction. Then the cross-modal contrastive learning can be formulated as:

\begin{equation}\label{eq:e2}\small
\begin{split}
&\ell_{itc}  = \frac{1}{2}\EE_{(\iii,\w)}[CE(\y^{i2t}(\iii),\p^{i2t}(\iii)) + CE(\y^{t2i}(\w),\p^{t2i}(\w))] \\ 
&  p_{b}^{i2t}(\iii)  = \frac{exp(s(\iii,\w_b)/\tau)}{\sum_{b=1}^B exp(s(\iii,\w_b)/\tau)}, p_b^{t2i}(\w)  = \frac{exp(s(\w,\iii_b)/\tau)}{\sum_{b=1}^B exp(s(\w,\iii_b)/\tau)}
\end{split}
\end{equation}
where $p_b^{i2t}(\iii)$ and $p_b^{t2i}(\w)$ denote softmax-normalized image-to-text and text-to-image similarity with batch size $B$ and temperature scale parameter $\tau$. $CE$ denotes cross-entropy loss. We follow ALBEF~\cite{albef} and use the momentum model to generate pseudo-targets as additional supervision.

\subsection{Similarity-bucket}

\begin{figure}[t]
	\centering
	\includegraphics[width=\linewidth]{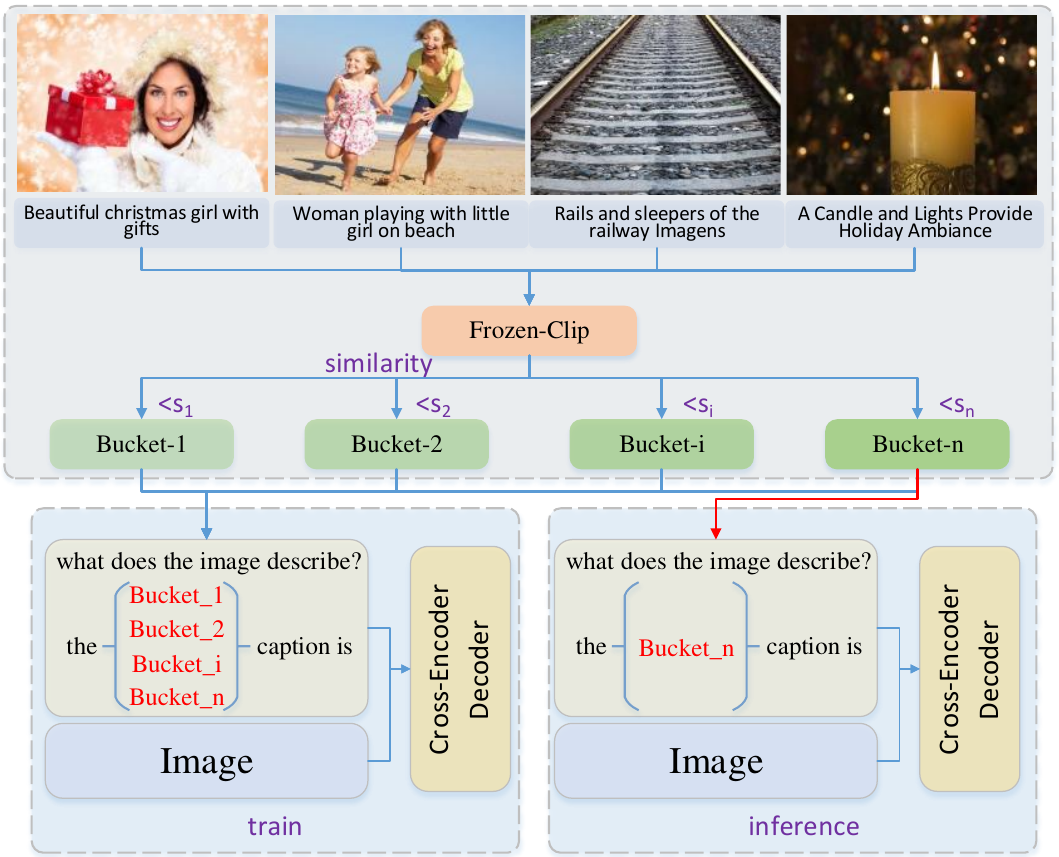}
	\caption{Similarity-bucket is utilized in pre-training, coarse-tuning, and fine-tuning stages. }\label{fig:sim-bucket}
\end{figure}

\textbf{Similarity-bucket} strategy provides different similarity prompts to the vision-language model in the pre-training, coarse-tuning, and fine-tuning stages. During the training part, we define \textit{n} buckets according to the similarity of training datasets in each stage. The similarity of the image-text pair in the first bucket is very low and may be noisy data. The \textit{$n{-}th$} bucket indicates that the similarity and quality of the image-text pair are very high. Through the similarity prompt, the model can learn the representation of training data with different qualities independently. In the inference part, the similarity-prompt is fixed to the \textit{n\textcolor{red}{-}th} bucket, forcing the model to generate higher quality and more matching captions.

As shown in figure \ref{fig:sim-bucket}, for example, in the pre-training stage, for each image-text pair, we utilize Frozen-Clip~\cite{clip} to predict the similarity of the image-text pair, and then we can obtain a similarity collection, which includes the similarity of all training data. With \textit{n} denoting the number of buckets, then based on the size of the dataset, we divide different similarity thresholds to represent each bucket. The result of splitting buckets satisfies the condition that the number of image-text pairs for the first and \textit{$n{-}th$} buckets is less than the number of image-text pairs for the remaining buckets.

In the training part, each image-text pair belongs to only one bucket, therefore, each image-text pair corresponds to a certain similarity prompt. We insert this similarity prompt into the template of the OFA model, resulting in the template: ``What does the image describe? The \{bucket\_i\} caption is''. Through this similarity prompt, the model can learn the representation of data with different qualities. In the inference part, we fix the similarity-prompt to the \textit{n-th} bucket and control the model to generate the most matched and high-quality caption for a given image.

\subsection{Retrieval-augmented}

\textbf{Retrieval-augmented}~\cite{retrieval-augmented} strategy provides a mini knowledge-base for each image-text pair during the training part. The model can not only extract visual features such as objects, attributes, and relationships of the image but also explicitly align the information of the image with the knowledge in the knowledge base.

\begin{figure}[t]
	\centering
	\includegraphics[width=\linewidth]{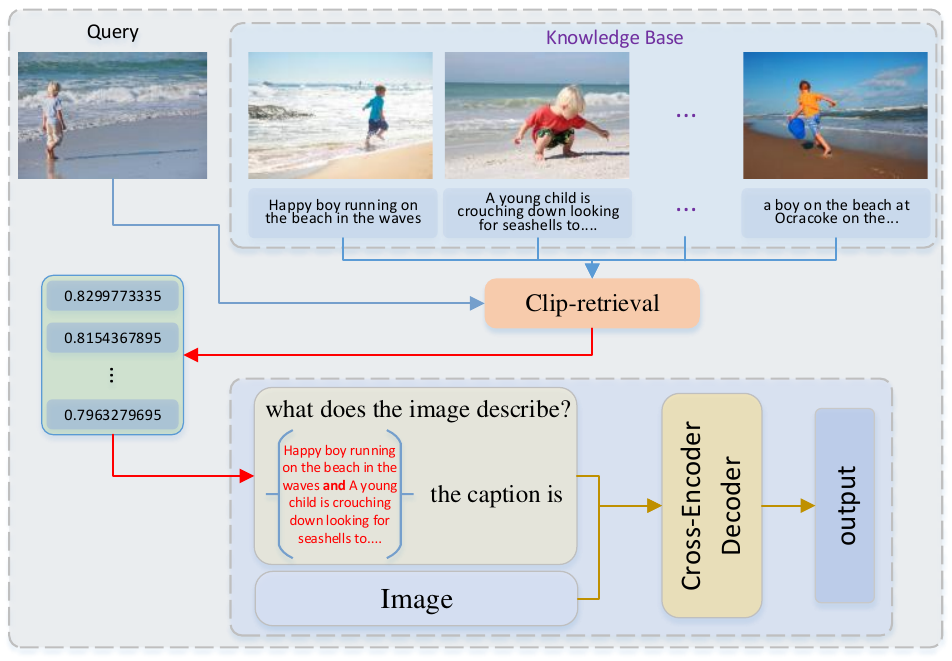}
	\caption{Similarity-bucket is utilized in pre-training, coarse-tuning, and fine-tuning stages. }\label{fig:retrieval-aug}
\end{figure}

As shown in figure \ref{fig:retrieval-aug}, for example, in the coarse-tuning stage, for each image query, we utilize \textit{Clip-retrieval}~\cite{clip-retrieval} library to retrieve \textit{top-k} image-text pairs according to the similarity between image query and each image of the Laion-5B dataset, and then we extract \textit{k} texts from the \textit{top-k} retrieved image-text pairs. These \textit{k} texts are concatenated into a mini knowledge-base and inserted into the template of the OFA model, resulting in the template: ``What does the image describe? \{retrieval knowledge\}, the caption is''. Through this retrieved knowledge, the model can generate content-rich and diverse captions.

\subsection{Model-ensemble}

The last stage is \textbf{model-ensemble}, we use similarity copy-paste and cider-ensemble tricks. Similarly copy-paste trick directly copies and pastes the caption through a fast shortcut. Specifically, in the training dataset, we compute the similarity between test data and training data. Then, we set a larger similarity threshold to select the most relevant image-text pair as the candidate pair. Finally, we define the similarity between the candidate caption and the candidate image as \textit{$c_1$}, and the similarity between the model prediction and the candidate image as \textit{$c_2$}, respectively, The caption corresponding to \textit{max($c_1, c_2$)} is used as the final prediction result. Cider-ensemble trick aims to find the best prediction result. Specifically, we fine-tune \textit{n} models to obtain \textit{n} prediction results, each prediction is calculated as a \textit{cider-score} along with other n-1 predictions, and the caption corresponding to the highest score is used as the final prediction result.

\section{Experiments}
\subsection{Implementation Detail}

\textbf{Dataset}. The training data at all stages are collected from \textit{Laion-5B}, a large-scale CLIP-filtered image-text dataset. Each image-text pair in \textit{Laion-5B} includes an image, a text, a URL, and the similarity between the image and text. In the pre-training stage, we collect \textit{6M} image-text pairs based on specific \textit{url} (\textit{thumbx.shutterstock.com, editorial01.shutterstock.com, etc.}), and extract \textit{top-1M} image-text pairs based on similarity. In the coarse-tuning stage, we use all the competition images to retrieve external data from \textit{Laion-5B} through \textit{Clip-retrieval} library. For each image query, we retrieve \textit{top-30} image-text pair based on the similarity between the image query and \textit{Laion-5B}, and retain \textit{120k} image-text pairs through simple data cleaning, such as filtering out too long, too short and non-English image-text pair. In the fine-tuning stage,  we also use all the competition images to retrieve external data from \textit{Laion-5B} through \textit{Clip-retrieval} library. For each image query, we retrieve \textit{top-10} image-text pair with the only URL \textit{www.tscdn.net} and retain \textit{12k} image-text pairs. In addition, we also add the \textit{5k} validation dataset to the fine-tuning stage.

\textbf{Model}. The base image captioning model we used is OFA, which is a large-scale visual-language pre-training model based on handcrafted templates. In each training stage, the number of similarity buckets is \textit{4}, and these four different similarity-prompts are \textit{``noise'', ``low quality'', ``high quality'', ``best match''}. In the coarse-tuning and fine-tuning stages, the numbers of retrieved relevant image-text pairs are \textit{1,2 and 4} so that we can obtain different models to execute the model-ensemble trick.

\textbf{Pre-training stage}. The size of the pre-training dataset is \textit{1M}, and we load the pre-trained ofa-large weights. Due to the large scale of the pre-training dataset, we only continued to pre-train \textit{5} epochs. We use \textit{NVIDIA RTX 3090$\times$4} with an initial learning rate of \textit{1e-5}, the input image size is \textit{380$\times$380}, the batch size is \textit{16}, and the input text length is \textit{30}. All other hyper-parameters follow the default setting of the OFA model.

\textbf{Coarse-tuning stage}. The size of the coarse-tuning dataset is \textit{120k}, For each image query, we retrieve \textit{top-30} image-text pairs based on similarity between the image query and \textit{Laion-5B}. We load the pre-trained weight from the pre-training stage and coarse-tune \textit{20} epochs. We use \textit{NVIDIA RTX 3090$\times$4} with an initial learning rate of \textit{1e-5}, the input image size is \textit{480$\times$480}, the batch size is \textit{16}, and the input text length is \textit{30}. All other hyper-parameters follow the default setting of the OFA model.

\textbf{Fine-tuning stage}. The size of the fine-tuning dataset is \textit{17k}, which includes \textit{12k} image-text pairs retrieved from \textit{Laion-5B} and \textit{5k} competition validation dataset. We load the coarse-tuned weight from the coarse-tuning stage and fine-tune \textit{100} epochs. We use \textit{NVIDIA RTX 3090$\times$4} with an initial learning rate of \textit{1e-5}, the input image size is \textit{480$\times$480}, the batch size is \textit{16}, and the input text length is \textit{30}. All other hyper-parameters follow the default setting of the OFA model.

\textbf{Model-ensenble stage}. We perform two random image augmentations for the test set and training set, obtaining \textit{4} similarities, and taking the average as the final similarity. The range of similarity values is between \textit{0.15$\sim$0.4}, and we set a large similarity threshold to \textit{0.35} to perform the copy-paste trick. By setting the number of different \textit{top-k} retrieval image-text pairs, taking weights for different epochs, and so on, we ultimately fuse \textit{20} models using the cider-score trick to obtain the final score.

\subsection{Result}

\begin{table}[!ht]
\centering
\caption{We report the Cider score of our methods on the test set.}
\begin{tabular}{ccc}% 其中，tabular是表格内容的环境；c表示centering，即文本格式居中；c的个数代表列的个数
\toprule[1.5pt]
\# & \textbf{Method}  &  \textbf{Cider} \\ %换行
\midrule %[2pt]  
1 & OFA+Pre-training,Coarse-tuning,Fine-tuning & 170+ \\
2 & Retrieval-augmented & 290+ \\
3 & Similarity-bucket & 310+ \\
4 & Model-ensemble & 325+ \\
\bottomrule[1.5pt]
\label{table-1}
\end{tabular}
\end{table}

As reported in table \ref{table-1}, with pre-training in the 1M dataset and fine-tuning in the 5k validation dataset, the performance of the OFA model reached a 170+ cider score. The most significant improvement strategy is the retrieval-augmented, which improves the cider score by 120+ points. By utilizing the similarity-bucket strategy and model-ensemble trick, we achieved the highest cider score on the leaderboard of 325+.

\section{Conclusion} This report summarizes our solution for the New frontiers for the Zero-shot Image Captioning challenge, which includes four core components: Pre-training, Coarse-tuning, Fine-tuning, and Model-ensemble. Our solution indicates that the similarity bucket is an effective paradigm for controlling the model to generate high-quality results for vision-language downstream tasks. The final competition results show the effectiveness of our solution.

%%%%%%%%% REFERENCES
{\small
\bibliographystyle{ieee_fullname}
\bibliography{output}
}

\end{document}